\documentclass{article}

\PassOptionsToPackage{numbers, compress}{natbib}
\usepackage[main, final]{neurips_2026}

\usepackage[utf8]{inputenc}
\usepackage[T1]{fontenc}
\usepackage{hyperref}
\usepackage{url}
\usepackage{booktabs}
\usepackage{amsfonts}
\usepackage{nicefrac}
\usepackage{microtype}
\usepackage{xcolor}
\usepackage{graphicx}
\usepackage{subcaption}
\usepackage{tcolorbox}
\usepackage{enumitem}
\usepackage{multirow}
\usepackage{pifont}
\usepackage{textcomp}
\usepackage{amsmath}
\usepackage{amssymb}
\usepackage{mathtools}
\usepackage{amsthm}
\usepackage[capitalize,noabbrev]{cleveref}
\usepackage[textsize=tiny]{todonotes}
\usepackage{algorithm}
\usepackage{algorithmic}
\usepackage{wrapfig}

\theoremstyle{plain}

\theoremstyle{definition}

\theoremstyle{remark}

\title{The Kernel Reality Check: Benchmarking and Distilling Efficient Attention at Scale}

\author{%
  F{\i}rat \"Oncel$^{1,2}$ \quad
  Cem Subakan$^{1,2,3}$ \quad
  Mirco Ravanelli$^{1,2}$ \quad
  \c{C}a\u{g}atay Y{\i}ld{\i}z$^{4}$ \\[4pt]
  $^1$Concordia University \quad
  $^2$Mila -- Quebec AI Institute \\
  $^3$Laval University \quad
  $^4$University of T\"ubingen \\[4pt]
  \texttt{firat.oncel@mila.quebec} \quad
  \texttt{cem.subakan@ift.ulaval.ca} \\
  \texttt{mirco.ravanelli@concordia.ca} \quad
  \texttt{cagatay.yildiz@uni-tuebingen.de}
}

\begin{document}
\maketitle

\begin{abstract}
Kernelized attention methods, which aim to replace the $O(N^2)$ complexity of softmax attention with $O(N)$ alternatives, have been the focus of intense research in the last five years. Despite growing interest, systematic evaluations and fair comparisons across approaches and scales are still missing. We analyze kernelized attention in two halves and highlight prominent limitations on each.
On efficiency, FlashAttention achieves lower memory consumption \emph{and} faster wall-clock time than SOTA kernelized methods during training and in vision transformers across most sequence lengths (with inference-time memory in causal language models the only clear exception), contradicting the core assumption for pure kernelized attention used as a drop-in replacement for softmax. We also discover that reported efficiency gains often fail to reproduce against modern baselines. On quality, current kernelized methods leave gaps of $2$-$10$ points relative to softmax under a unified distillation protocol, and model scala alone does not close these gaps.
We show this gap is not intrinsic to kernelization: LARA++, a principled extension of LARA~\citep{zheng2022linear} that approximates softmax via importance sampling with adaptive proposal distributions, recovers up to 99\% of softmax accuracy. We establish that the gap is closable; however, the computational overhead remains substantial, leaving the efficiency wall intact. Code is available at\footnote{\url{https://github.com/firatoncel/KernelRealityCheckNeurIPS2026}}
\end{abstract}

\section{Introduction}

Transformer networks have become the dominant architecture across machine learning, yet their $O(N^2)$ attention complexity poses a fundamental scalability challenge. A principled line of research has sought to address this through kernelization of the attention mechanism, replacing the query-key outer product with feature maps that yield theoretically elegant $O(N)$ space and compute complexity. Over the past five years, this approach has produced increasingly sophisticated methods: randomized feature maps~\citep{choromanski2020rethinking}, learned projections~\citep{zhang2024hedgehog}, importance sampling schemes~\citep{zheng2022linear}, and frequency-domain transformations~\citep{chen2024DiJiang}. Each promises substantial memory reductions and computational speedups. However, despite these theoretical advances, kernelized attention has not displaced standard transformers in practice. FlashAttention~\citep{dao2023flashattention2, dao2022flashattention}, which computes \emph{exact} softmax attention through hardware-aware implementation with $O(N)$ space and $O(N^2)$ computation complexity, dominates production deployments. Given these observations, we ask:

\begin{tcolorbox}[colback=gray!10,colframe=black,boxrule=0.5pt,arc=0pt,left=3pt,right=3pt,top=3pt,bottom=3pt,width=\linewidth]
\centering
Do the theoretical complexity gains of kernelized attention translate to practical efficiency?
\end{tcolorbox}

We conduct a systematic investigation across model scales and find that, largely, they do not. We evaluate across two domains: causal language models (CLMs), an important testbed for kernelized attention, and vision transformers (ViTs), where these approaches have received limited systematic investigation. Our evaluation proceeds in two halves.
On the \emph{efficiency} axis, FlashAttention achieves lower memory consumption than highly optimized kernelized methods across most sequence lengths and matches or beats them on wall-clock time except at very long sequences ($\geq 32$K). The clear wins for linear attention are confined to inference in CLMs: memory during prefill and auto-regressive generation. These findings expose the gap between asymptotic analysis and practical performance: the $O(N)$ complexity advantage of linear attention often does not translate to wall-clock or memory wins against hardware-aware exact methods.
On the \emph{quality} axis, SOTA kernelized methods fail to recover softmax accuracy under a unified distillation protocol: Hedgehog~\citep{zhang2024hedgehog} and DiJiang~\citep{chen2024DiJiang} leave gaps of $2$--$10$ points across DeiT configurations (classification), and originally reported gains do not consistently reproduce. To test whether this quality gap is intrinsic to kernelization or specific to prior designs, we develop LARA++, a principled extension of the importance-sampling formulation of LARA~\citep{zheng2022linear}. LARA++ recovers up to $99\%$ of softmax accuracy on vision transformer distillation, demonstrating that the gap is closable. Nevertheless, it inherits LARA's substantial computational overhead, leaving the efficiency wall of the first half intact.

Our study makes the following contributions:
\begin{itemize}[leftmargin=*,itemsep=1pt,topsep=2pt]
\item \textbf{A unified benchmarking framework} for kernelized attention measuring actual memory and throughput, addressing the fragmented evaluation landscape that has obscured comparisons.
    \item \textbf{Systematic reproduction and comparison} of SOTA methods under consistent experimental conditions, revealing significant discrepancies with published results.
    \item \textbf{LARA++}, a principled extension of LARA's importance-sampling formulation, closes the quality gap left by prior kernelized methods. This shows that the gaps left by prior methods are not intrinsic to kernelization, and points to computational cost as the binding constraint.
\end{itemize}

Taken together, our findings question whether pure kernelized attention, used as a drop-in replacement for softmax in transformers, offers practical advantages over hardware-optimized exact methods: the speed–memory trade-offs do not materialize even when approximation quality is addressed. The success of hybrid approaches~\citep{hua2022transformer, zhang2024lolcats}, including BASED~\citep{arora2024simple} and GLA~\citep{yang2023gated}, which report prefill and training-throughput gains over FlashAttention, points to a more viable path: combining selective linear attention with exact local computation, where the two compose rather than compete.

\vspace{-1.0em}
\section{Background on kernelized attention}\vspace{-0.25em}
\label{sec:kernelized}
\vspace{-0.5em}

We begin by reviewing the kernelization framework for self-attention, which provides the theoretical foundation for linear complexity attention mechanisms. We then examine the computational challenges that arise under causal masking. For clarity, we present single-head formulations throughout; extension to multi-head attention is straightforward.

Consider an input sequence $\mathbf{X} \in \mathbb{R}^{N \times d}$ where $N$ denotes sequence length and $d$ the feature dimension. Standard self-attention computes queries, keys, and values via learned projections:
\begin{align}
\mathbf{Q} &= \mathbf{X}\mathbf{W}_Q, \quad \mathbf{K} = \mathbf{X}\mathbf{W}_K, \quad \mathbf{V} = \mathbf{X}\mathbf{W}_V
\end{align}
where $\mathbf{W}_Q, \mathbf{W}_K \in \mathbb{R}^{d \times d_k}$ and $\mathbf{W}_V \in \mathbb{R}^{d \times d_v}$. The output for query position $i$ is computed as:
\begin{equation}
\mathbf{y}_i = \sum_{j=1}^{N} \frac{\exp(\mathbf{q}_i^T \mathbf{k}_j / \sqrt{d_k})}{\sum_{l=1}^{N} \exp(\mathbf{q}_i^T \mathbf{k}_l / \sqrt{d_k})} \mathbf{v}_j
\end{equation}
where $\mathbf{q}_i, \mathbf{k}_j \in \mathbb{R}^{d_k}$ denote rows of $\mathbf{Q}, \mathbf{K}$ and $\mathbf{v}_j \in \mathbb{R}^{d_v}$ denotes the $j$-th row of $\mathbf{V}$. This formulation incurs $O(N^2 d_k)$ time complexity and $O(N^2)$ memory for the attention matrix.

Absorbing the scaling factor into the projections, the attention weights can be expressed via an exponential kernel $k(\mathbf{q}_i, \mathbf{k}_j) = \exp(\mathbf{q}_i^T \mathbf{k}_j)$, yielding:
\begin{equation}
\mathbf{y}_i = \frac{\sum_{j=1}^{N} k(\mathbf{q}_i, \mathbf{k}_j) \mathbf{v}_j}{\sum_{j=1}^{N} k(\mathbf{q}_i, \mathbf{k}_j)}
\end{equation}

The kernel trick enables us to express this kernel via a feature map $\phi: \mathbb{R}^{d_k} \rightarrow \mathbb{R}^{d'}$ such that $k(\mathbf{q}, \mathbf{k}) = \phi(\mathbf{q})^T \phi(\mathbf{k})$. This reformulation permits rearranging the computation order:
\begin{equation}
\label{eq:linear-attention}
\mathbf{y}_i = \frac{\phi(\mathbf{q}_i)^T \mathbf{S}}{\phi(\mathbf{q}_i)^T \mathbf{z}}, ~~ \text{with} ~~ \mathbf{S} = \sum_{j=1}^{N} \phi(\mathbf{k}_j) \mathbf{v}_j^T, ~~ \mathbf{z} = \sum_{j=1}^{N} \phi(\mathbf{k}_j)
\end{equation}

Crucially, $\mathbf{S} \in \mathbb{R}^{d' \times d_v}$ and $\mathbf{z} \in \mathbb{R}^{d'}$ can be computed once and reused for all queries, reducing complexity to $O(Nd'd_v)$ time and $O(d'd_v)$ memory when $d' \ll N$. 
\citep{kasai2021finetuning} further reformulated the attention mechanism to process tokens sequentially, maintaining cumulative states that are updated at each timestep. Specifically, at position $i$, we maintain:
\begin{align}
\mathbf{S}_i &= \mathbf{S}_{i-1} + \phi(\mathbf{k}_i) \mathbf{v}_i^T, \qquad
\mathbf{z}_i = \mathbf{z}_{i-1} + \phi(\mathbf{k}_i)
\end{align}
with initial conditions $\mathbf{S}_0 = \mathbf{0}$ and $\mathbf{z}_0 = \mathbf{0}$. The output at position $i$ is then computed as $\mathbf{y}_i = \phi(\mathbf{q}_i)^T \mathbf{S}_i / (\phi(\mathbf{q}_i)^T \mathbf{z}_i)$. This view establishes a direct connection between kernelized attention and recurrent neural networks, enabling efficient inference with constant memory footprint per timestep.

\textbf{Causal masking challenge. }
While the kernelization framework elegantly reduces complexity in the bidirectional setting, causal attention introduces fundamental computational difficulties during training. In non-causal attention, a single cache $\mathbf{S} = \sum_{j=1}^{N} \phi(\mathbf{k}_j) \mathbf{v}_j^T$ suffices for all queries, requiring only $O(d'^2)$ memory and constant computation per new input. However, as noted by \citep{hua2022transformer}, causal constraints force the model to compute and maintain $N$ distinct cache values $\{\mathbf{S}_i\}_{i=1}^{N}$, one for each query position, since each $\mathbf{S}_i$ depends on a different prefix of the sequence.

While theoretically these states can be computed in $O(Nd'^2)$ time via cumulative summation, this introduces an RNN-style sequential dependency across all $N$ timesteps. This sequential dependency severely limits parallelism and necessitates $N$ memory accesses within the computation loop. On modern accelerators, memory access latency typically dominates arithmetic costs, creating a substantial gap between theoretical complexity and actual wall-clock time. Indeed, our findings in Section~\ref{sec:efficiency} show that hardware-aware exact attention outpaces optimized linear-attention kernels in this regime, despite the latter's theoretical advantage.

\section{Related work}\vspace{-0.5em}
Improving transformer efficiency has been a central research challenge, and kernelized attention has attracted substantial interest due to its theoretical elegance. Yet despite the algorithmic advances in the last six years, these methods have seen limited production adoption. We organize related work into three categories: \textit{kernelized attention} approaches that reformulate attention through kernel theory, \textit{hybrid methods} that augment kernelization with exact local attention, and \textit{implementation-level} optimizations that we view as establishing practical efficiency bounds.

\vspace{-0.5em}
\paragraph{Kernelized attention approaches}
\citep{katharopoulos2020transformers} proposed the earliest work on improving efficiency, where self-attention is reformulated as a linear dot-product of kernel feature maps with \texttt{elu} as the feature function. While effective, this feature map is arbitrary. Most subsequent works, instead, design feature maps based on kernel methods~\citep{bochner2005harmonic, rahimi2007random}. \citep{peng2021random} replace the softmax kernel $\exp(\langle\cdot,\cdot\rangle)$ with its \textit{unbiased} estimate using random features sampled from a Gaussian distribution with zero mean and fixed variance (the \textit{spectral density}). In closely related work, \citep{choromanski2020rethinking} introduce positive orthogonal random features (PERF), which replace standard random features with orthogonal ones and provably approximate the softmax kernel. \citep{chowdhury2022learning} propose \textit{learning} the spectral distribution via a mixture of Gaussians or generative model. \citep{zhang2024hedgehog} argue that existing approaches cannot capture the spiky, concentrated nature of softmax-induced distributions, instead propose learning a one-layer MLP that predicts features deterministically and thereby departing from the randomized attention framework.

Beyond feature map design, more sophisticated approaches reformulate the underlying computation. \citep{chen2024DiJiang} improve sample efficiency through weighted Quasi-Monte Carlo integration (replacing standard Monte Carlo sampling with respect to the spectral density) and accelerate computation via frequency domain transformations. \citep{zheng2022linear} approximate the same integration using importance sampling, where the proposal distribution adapts to queries and keys. \citep{chen2023primal} show that self-attention output solves a constrained optimization problem and propose computing the primal solution directly. Notably, each generation of methods introduces additional complexity; learned feature maps, frequency transformations, optimization-based reformulations.

\paragraph{Hybrid methods: combining kernelization with exact attention}
Pure kernelized approaches have struggled to match softmax quality, motivating hybrid architectures that combine linear attention with other ideas. For example, unlike the common practice of training from scratch, \citep{kasai2021finetuning} adopt an alternative strategy: replacing self-attention layers in a \textit{pretrained} network with linear counterparts, then finetuning using the original objective.
cosFormer~\citep{qin2022cosformer} replaces softmax with ReLU and introduces a distance reweighting mechanism to stabilize training and concentrate attention on local tokens.
\citep{hua2022transformer} propose FLASH, introducing a gated attention unit (GAU) that enables effective attention combined with mixed attention, where they apply \textit{quadratic attention locally} within chunks and linear attention globally across chunks to achieve transformer-quality performance with linear complexity. 

Recent methods increasingly combine linear and exact local attention. \textsc{LoLCATS}~\citep{zhang2024lolcats} uses a local window with attention transfer following \citep{kasai2021finetuning}, while \citep{zheng2023efficient} reduce random-feature variance with control variates that also require local attention. Lightning Attention~\citep{qin2024various} addresses causal masking via intra-block quadratic and inter-block linear attention. BASED~\citep{arora2024simple} combines a Taylor feature map, sliding-window attention, and short convolutions, reporting $56\%$ faster prefill than FlashAttention-2 at $N=4$K for a 1.3B model. GLA~\citep{yang2023gated} adds data-dependent gating with chunkwise training and reports higher throughput beyond length 4096. Both BASED and GLA train from scratch rather than replacing attention in pretrained models.

The trajectory of this subfield reveals that the most successful methods increasingly rely on exact attention components, raising the question of how much benefit kernelization itself provides.

\vspace{-0.5em}
\paragraph{Implementation-level optimizations}
A parallel research direction sidesteps algorithmic reformulation, instead accelerating \textit{exact} self-attention through hardware-aware implementation. These methods have achieved remarkable practical success and represent the de facto efficiency baseline in production systems. FlashAttention~\citep{dao2022flashattention} uses IO-aware tiling to reduce HBM--SRAM transfers and avoid materializing the full $O(N^2)$ attention matrix, improving speed and memory use. FlashAttention-2~\citep{dao2023flashattention2} and FlashAttention-3~\citep{shah2024flashattention3} add further GPU-level optimizations. On the kernelized side, FlashLinearAttention~\citep{yang2024fla}, used by GLA~\citep{yang2023gated}, provides Triton-optimized chunked prefix-sum kernels for parallel training. RingAttention~\citep{liu2023ring} distributes attention across devices for longer sequences, while PagedAttention~\citep{kwon2023efficient} reduces KV-cache fragmentation through non-contiguous paging.

\vspace{-0.5em}
\paragraph{Limitations of existing evaluations}
Our literature review reveals systematic evaluation gaps that obscure the true state of the field:
First, SOTA kernelized methods are rarely compared against one another or against FlashAttention. SOTA \citep{zhang2024hedgehog} and \citep{chen2024DiJiang} evaluate against methods predating 2022, while later work~\citep{qin2024various, zhang2024lolcats} omits comparisons with these recent approaches. Second, efficiency claims rely on narrow setups with underspecified details and poorly justified baselines. Third, evaluation protocols lack consistency: benchmarked tasks, model scales, metrics, and settings (training from scratch versus converting pretrained models) vary substantially across studies, preventing direct comparison. For example, \citep{zhang2024hedgehog} report associative recall performance of BERT \citep{devlin-etal-2019-bert}, Rouge score on Llama2 \citep{touvron2023llama2}, standard results on LRA \citep{lra}, perplexity on GPT-2 \citep{radford2019language}; \citep{chen2024DiJiang} report downstream task comparisons against Pythia \citep{biderman2023pythia}, Mamba \citep{mamba}, and Llama2; \citep{zheng2022linear} evaluate on classification, video action recognition, and machine translation. These gaps allow contradictory claims to persist unchallenged. To address these limitations, we propose a unified framework that systematically evaluates the memory consumption and execution time of SOTA kernelized attention method (\cref{sec:efficiency}).

\vspace{-0.5em}
\section{Memory and computational efficiency analysis of kernelized attention}
\label{sec:efficiency}
\vspace{-0.5em}

To test whether kernelized attention methods actually have practical advantages, we profile actual memory consumption and computational throughput of the following attention approaches:
\vspace*{-.1cm}
\begin{itemize}[leftmargin=15pt]
    \item \textbf{Softmax attention.} Standard attention serves as our performance reference, establishing the accuracy target kernelized methods aim to recover. We use the naive PyTorch  \citep{pytorch} implementation.
    \item \textbf{FlashAttention.} FlashAttention~\citep{dao2023flashattention2, dao2022flashattention} computes exact softmax through IO-aware tiling. As the dominant attention implementation in production systems, it establishes the practical efficiency baseline that kernelized methods must beat to justify their approximation error. We use FlashAttention v2.3.6~\citep{dao2023flashattention2}.
    \item \textbf{Linear attention.} We use Hedgehog~\citep{zhang2024hedgehog} and DiJiang~\citep{chen2024DiJiang}, both representing the current state-of-the-art in kernelized attention and reporting substantial efficiency gains over exact methods. Hedgehog learns feature maps via MLPs to capture spiky softmax distributions. DiJiang employs frequency domain transformations, and introduces learnable scaling parameters and a gating mechanism. For fair comparisons against well-engineered FlashAttention, we train these models using Triton-optimized chunked parallel prefix-sum implementation of FlashLinearAttention \citep{yang2024fla}. During generation, we employ Triton-fused $O(1)$ recurrent kernels for both methods to ensure comparable optimization levels to FlashAttention's CUDA implementation. 
\end{itemize}

\paragraph{Experimental setup.}
We measure memory consumption and computational throughput across two 
architectures and two operational phases. For \textit{vision transformers},  we profile ViT-S, ViT-B, and ViT-L on an A100 GPU across image resolutions from 224px to 1536px, corresponding to token sequence lengths $N \in [197, 9217]$,  reporting inference and training time and memory. For \textit{causal language  models}, we profile GPT2-S, GPT2-M, and GPT2-L. We separate two phases: \textit{prefill} (a single forward pass encoding all input tokens) and \textit{auto-regressive 
generation}, where the model produces one new token per step conditioned on  all preceding context (we do not benchmark training-time generation, as it involves no token-by-token decoding). All measurements are averaged over 100 runs. All methods share the same harness, host model, dtype, and batch size; timing follows a warm-up phase and uses CUDA-synchronised wall-clock time, and peak memory is read from \texttt{torch.cuda.max\_memory\_allocated}.
\vspace{-0.5em}

\begin{figure*}
    \centering
    \includegraphics[width=0.99\linewidth]{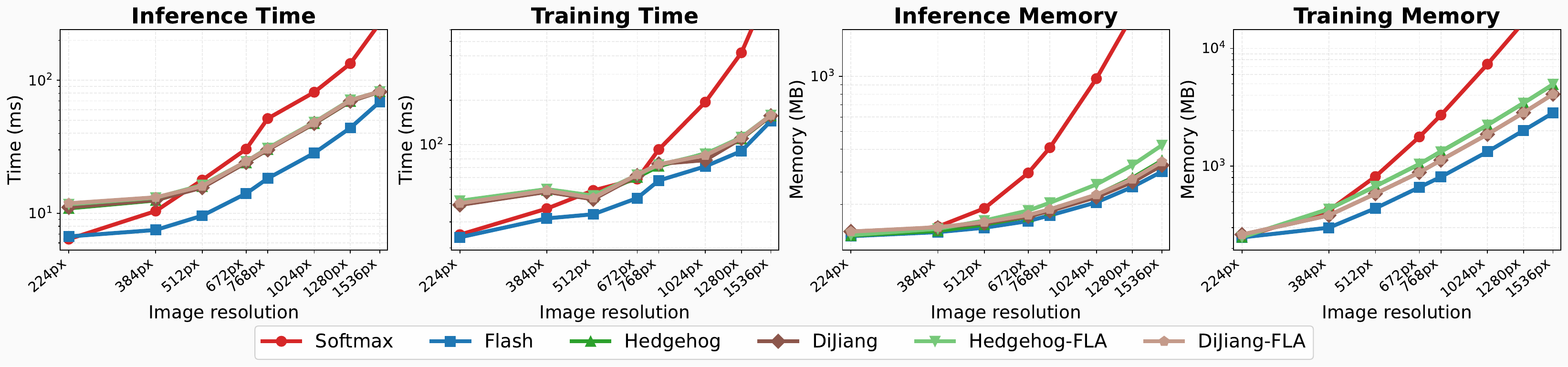}
    \caption{Memory and computational efficiency across sequence lengths (ViT-S, A100 GPU).}
    \label{fig:efficiency}
\end{figure*}

\paragraph{Finding 1: FlashAttention is more efficient than pure kernelized attention during training.}
Across both vision and language model settings, FlashAttention outperforms all kernelized methods in training time and memory. On ViT-S at 1536px, it uses 2{,}830MB versus 4{,}085MB for DiJiang, a 1.4$\times$ memory reduction, while also being faster (144.9ms vs.\ 157ms). The same pattern holds in causal language models: at $N=16$K, FlashAttention uses 11{,}767MB and 488.6ms, compared with 15{,}423MB and 565.8ms for DiJiang-FLA. These advantages persist across all tested scales (\cref{tab:scale-vit,tab:scale-gpt}). In vision, the memory advantage remains about 1.4$\times$ from ViT-S to ViT-L, while the speed gap narrows with model size without reversing. GPT2 shows the same trend: at $N=4$K, FlashAttention is about 1.1--1.2$\times$ faster and 1.3$\times$ more memory-efficient than the best kernelized alternatives. At $N=16$K on GPT2-L, Hedgehog-FLA runs out of memory while FlashAttention still fits. Thus, the theoretical $O(N)$ training complexity of kernelized attention does not translate into practical gains over a well-engineered exact implementation.
\vspace{-0.5em}
\begin{figure*}
    \centering
    \includegraphics[width=0.99\linewidth]{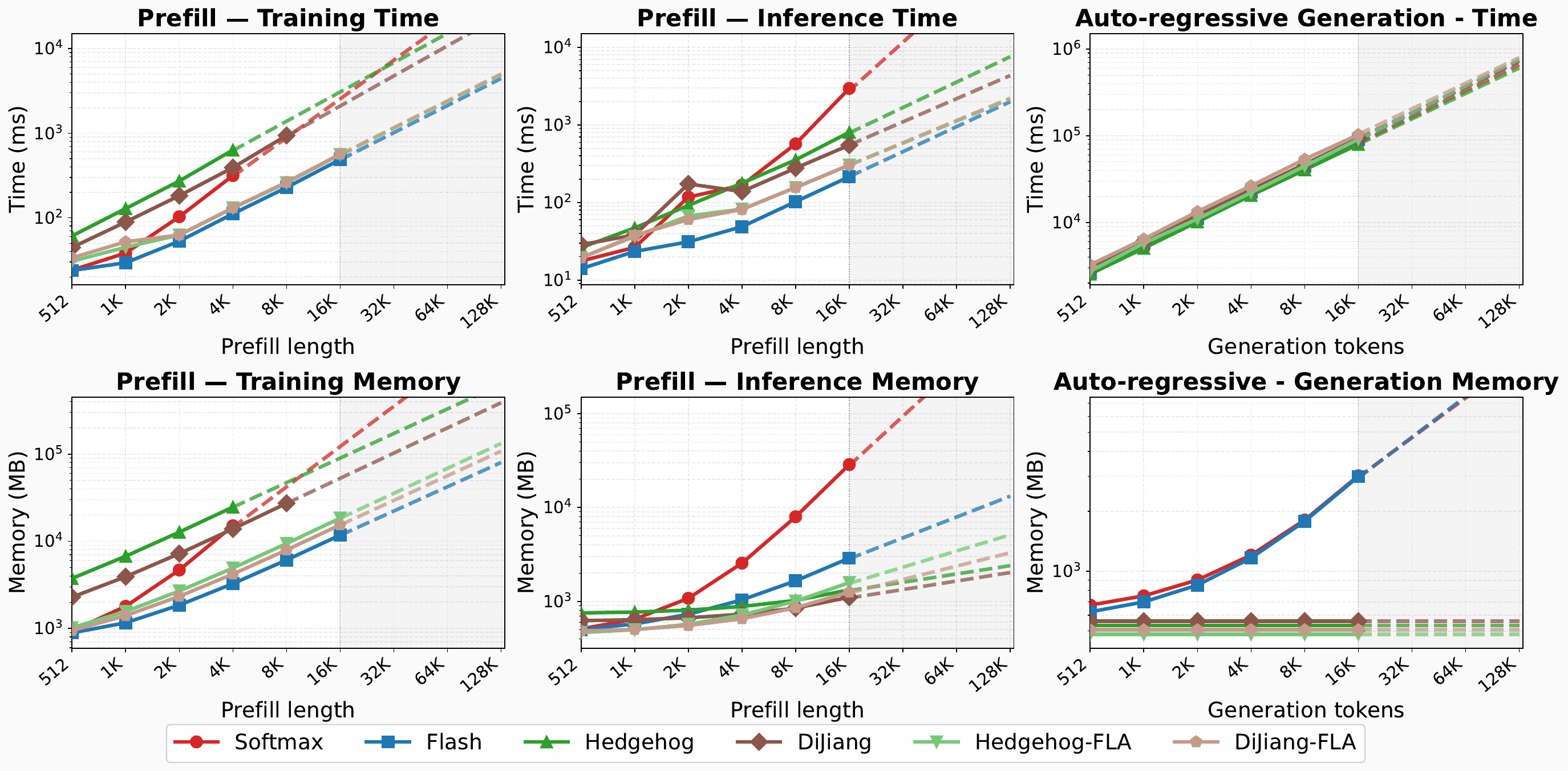}
    \caption{Memory and computational efficiency across sequence lengths (GPT2-S, A100 GPU). Markers and solid lines are measured; dashed lines are log-domain extrapolations beyond 16K tokens or past out-of-memory.}
    \label{fig:efficiency-causal}
\end{figure*}

\vspace{-.3em}
\paragraph{Finding 2: Linear attention offers no efficiency advantage in vision transformers.}
The bidirectional, non-causal structure of ViT is precisely where linear attention should be most favorable since there is no causal masking overhead, and the global key-value cache $\mathbf{S}$ can be shared across queries. Yet FlashAttention is faster \emph{and} more memory-efficient than kernelized methods at every resolution, in both inference and training. At 1536px, FlashAttention runs inference in 69.0ms using 302MB, while Hedgehog requires 82.5ms and 347MB; naive softmax consumes 4{,}294MB for the same pass. The ordering is unchanged on ViT-B and ViT-L: FlashAttention is the fastest and lowest-memory method in all six model/phase combinations of \cref{tab:scale-vit}, although the time margin shrinks with model size (inference: 1.19$\times$, 1.12$\times$, 1.07$\times$ over the fastest kernelized method for ViT-S/B/L).
\vspace{-0.5em}
\vspace{-.3em}
\paragraph{Finding 3: Auto-regressive generation and prefill memory are the clear wins for linear attention.}
In the recurrent formulation, linear attention maintains a fixed-size state throughout generation: Hedgehog and DiJiang consume ${\sim}$533MB and ${\sim}$560MB regardless of the number of tokens generated. On the contrary, FlashAttention must attend over a growing KV cache, climbing from 578MB at $N=128$ to 2{,}991MB at $N=16k$ generated tokens. FlashAttention's memory remains practically manageable (under 3GB at 16K tokens on an A100, a single input), so this memory advantage is more consequential in tight deployment constraints or at long generation tasks. The advantage persists at scale: after generating $16$K tokens, Hedgehog uses $5.6\times$, $5.8\times$, and $5.0\times$ less memory than FlashAttention on GPT2-S/M/L (\cref{tab:scale-gpt}); on GPT2-L, FlashAttention's KV cache reaches 15{,}576MB versus 3{,}144MB. Prefill inference memory shows the same pattern: at $N=16$K, DiJiang uses $2.6\times$, $3.1\times$, and $3.0\times$ less memory than FlashAttention, although FlashAttention remains the fastest prefill method at every scale. Causal inference, particularly long generation with larger models, is therefore the regime in which kernelized attention is preferable on efficiency grounds.
\vspace{-0.5em}
\vspace{-.3em}
\paragraph{Finding 4: Reported efficiency gains do not reproduce against modern baselines.}
Two representative claims from the literature fail under our evaluation protocol. First, \citet{chen2024DiJiang} report constant generation throughput as a distinguishing advantage of DiJiang (their Figure~4). We reproduce this: DiJiang maintains ${\sim}$174 tokens/s across $N \in [128, 16384]$. However, cached FlashAttention is similarly flat, at 193$\,\rightarrow\,$182 tokens/s over the same range. Although each decoding step still has $O(t)$ compute and $O(N)$ KV-cache memory, at these lengths latency is dominated by model-weight streaming, so throughput changes little. Comparing against a vanilla Transformer without KV cache is therefore misleading. The same pattern holds on GPT2-L, where both DiJiang and cached FlashAttention remain nearly flat.

Second, \citet{zhang2024hedgehog} report nearly $5\times$ faster inference than FlashAttention at $N=32$K (their Figure~6), using FlashAttention~v1~\citep{dao2022flashattention} on a single attention layer. Under our protocol (GPT2-S with FlashAttention~v2), we observe the opposite: Hedgehog prefill is $3.5\times$ \emph{slower} at $N=8{,}192$, and Hedgehog-FLA remains $1.4\times$ slower with Triton-optimized kernels. The reversal persists on GPT2-L, where Hedgehog is $2.4\times$ slower ($1.3\times$ with FLA). Both discrepancies reflect the evaluation gaps noted earlier: outdated baselines and narrow benchmarks that do not transfer to current settings.
\vspace{-0.5em}
\paragraph{Relation to GLA and BASED.}
Two hybrid architectures report regimes where they outperform FlashAttention-2: GLA~\citep{yang2023gated} in training throughput beyond $N=4$K, and BASED~\citep{arora2024simple} in prefill at $N=4$K. Our results reinforce the importance of hardware-aware implementation. Replacing naive kernels with FLA's Triton kernels speeds up Hedgehog prefill at $N=16$K by $2.6\times$, $2.1\times$, and $1.9\times$ on GPT2-S/M/L, closing $82$--$85\%$ of the gap to FlashAttention-2. A residual $1.2$--$1.4\times$ slowdown remains, however. Because we evaluate each kernelized method with both naive and FLA-optimized kernels, this separates much of the implementation effect from the underlying attention mechanism, although attributing the remaining gap would require a roofline analysis.

We also benchmark BASED using its released 360M checkpoints and CUDA kernel. Finding~1 carries over: FlashAttention uses $1.66\times$ less training memory across sequence lengths. BASED retains the generation-memory advantage of Finding~3, but not the prefill-memory advantage: FlashAttention uses $2.6\times$ less memory during prefill. GLA and BASED combine kernelized global attention with exact local computation and are trained from scratch, so they remain outside the drop-in setting considered by our main conclusions.

\paragraph{Summary.} Taken together, our results challenge the assumption that kernelized attention is more efficient than exact methods in practice: FlashAttention matches or surpasses all linear-attention variants in memory and speed across every setting except causal-LM inference, where kernelized methods retain advantages in prefill and generation memory and, for Hedgehog, generation latency. These exceptions persist or widen from GPT2-S to GPT2-L. Our measurements cover models up to 774M parameters (GPT2-L) and 304M parameters (ViT-L) on a single A100. We do not claim the ordering transfers unchanged to multi-billion-parameter models or other hardware, but none of the trends in \cref{tab:scale-vit,tab:scale-gpt} reverses across the model scales we test.
\vspace{-0.5em}
\paragraph{From efficiency to quality.}
The results above show that current kernelized methods fail to deliver the efficiency gains promised by their asymptotic analyses. A natural counter-argument is that efficiency may not be the only relevant axis, and kernelization could still be justified by quality preservation. We test this in \cref{sec:quality}, where SOTA methods fail to recover softmax accuracy under our distillation protocol. To determine whether this reflects specific designs (Hedgehog's MLPs, DiJiang's frequency transforms) or the kernelized framework itself, we develop LARA++ (\cref{sec:lara}), a principled extension of LARA's importance-sampling formulation. LARA++ recovers up to $99\%$ of softmax accuracy, showing that the quality gap is closable within the framework, but it inherits LARA's substantial computational overhead and does not escape the efficiency wall.

\vspace{-.2em}
\section{A unified benchmarking framework: distilling pretrained transformers}
\label{sec:benchmarking}
\vspace{-0.5em}
The efficient attention literature rests on a core assumption: replacing $O(N^2)$ softmax with $O(N)$ kernelized alternatives yields substantial savings. \cref{sec:efficiency} examined the efficiency half of this claim; we now turn to quality. Our framework prioritizes three principles: realistic model scales using production-grade architectures, consistent
evaluation protocols enabling direct comparison across methods, and comprehensive efficiency profiling.
\vspace{-0.5em}
\paragraph{Distillation-based protocol.}
Training foundation models from scratch requires computational resources and proprietary datasets unavailable to most academic groups. We instead adopt a distillation protocol: pretrained transformer attention layers are converted to kernelized variants, and we measure the resulting performance degradation. This setup reflects realistic deployment scenarios where practitioners compress existing models, holds the architecture and data fixed so that only the attention operator changes, and enables evaluation at scales otherwise infeasible.
\vspace{-0.5em}
\paragraph{Task selection.}
We evaluate on two complementary vision tasks. For \emph{supervised classification}, we distill DeiT~\citep{touvron2021deit} on ImageNet~\citep{imagenet} across three configurations (DeiT-S/224, DeiT-B/224, DeiT-B/384), spanning model depth and sequence lengths $N \in \{197, 577\}$. For \emph{self-supervised learning}, we distill DINOv2~\citep{oquab2024dinov} models (S, B, L, G, up to 1.1B parameters; $N=257$) using YFCC15M~\citep{yfcc15m}, since the original training corpus is unavailable. Distilled DINOv2 models are evaluated on ImageNet classification, VOC2012 (21-class)~\citep{everingham2010pascal}, and ADE20K (150-class)~\citep{zhou2017scene} semantic segmentation, testing
whether kernelized attention preserves representations that transfer beyond the distillation objective. Full hyperparameters are in \cref{app:exp-details}.
\vspace{-0.5em}
\section{SOTA kernelized methods fail to match softmax}
\label{sec:quality}
\vspace{-0.5em}
\begin{wraptable}[14]{r}{0.53\textwidth}
  \centering
  \vspace{-1.6em}
\caption{Top-1 ImageNet accuracy of distilled DeiT models. Linear attention methods consistently fail to recover softmax accuracy, leaving gaps of $1.8$--$10$ points. LARA++ closes the gap to within $1.3$ point.}
\small
\setlength{\tabcolsep}{4pt}
\begin{tabular}{l cc cc cc}
\toprule
& \multicolumn{2}{c}{DeiT-S/224} & \multicolumn{2}{c}{DeiT-B/224} & \multicolumn{2}{c}{DeiT-B/384} \\
\cmidrule(lr){2-3}\cmidrule(lr){4-5}\cmidrule(lr){6-7}
& Acc & $\Delta$ & Acc & $\Delta$ & Acc & $\Delta$ \\
\midrule
Softmax           & 79.8 & ---       & 81.8 & ---       & 82.9 & ---       \\
\midrule
\multicolumn{7}{l}{\emph{Linear attention}} \\
\quad Hedgehog    & 77.7 & $-2.1$    & 79.7 & $-2.1$    & 80.4 & $-2.5$    \\
\quad DiJiang     & 71.9 & $-7.9$    & 74.5 & $-7.3$    & 72.8 & $-10.1$   \\
\quad LARA        & 77.4 & $-2.4$    & 80.0 & $-1.8$    & 81.0 & $-1.9$    \\
\midrule
LARA++            & \textbf{78.5} & $-1.3$ & \textbf{80.8} & $-1.0$ & \textbf{82.1} & $-0.8$ \\
\bottomrule
\end{tabular}
\label{tab:deit-quality}
\end{wraptable}
We apply the protocol of \cref{sec:benchmarking} to test whether SOTA kernelized methods preserve softmax quality. We compare three representative approaches: \emph{Hedgehog}~\citep{zhang2024hedgehog}, which learns feature maps via small MLPs; \emph{DiJiang}~\citep{chen2024DiJiang}, which combines frequency-domain transforms with learned scaling and gating; and \emph{LARA}~\citep{zheng2022linear}, which approximates softmax via self-normalized importance sampling with adaptive proposals. 
\cref{tab:deit-quality} 
reports top-1 ImageNet accuracy after distillation across three DeiT configurations.

The linear attention group consistently underperforms softmax. Gaps range from modest (Hedgehog and LARA: $1.8$--$2.5$ points) to severe (DiJiang: $7.3$--$10.1$ points), and persist across model sizes and resolutions. Crucially, the gaps do not close as model capacity grows: DeiT-B/224 shows the same Hedgehog and LARA gaps as DeiT-S/224, and DiJiang's gap actually widens. We tuned each baseline's training recipe to its best stable configuration (\cref{app:details-compared-methods}); these are not numbers from undertrained baselines. The gaps are also not seed noise: repeating the DeiT-S runs with three independent seeds gives standard deviations of at most $0.15$ points for every baseline (\cref{tab:seeds}), far below the $2$--$8$ point gaps. Recipe differences nonetheless remain: DiJiang does not train under its published procedure, and Hedgehog needs a much smaller second-stage learning rate than published. These are reproducibility findings in their own right, and they mean the gaps cannot be attributed to attention expressiveness alone. LARA++ received less tuning than the baselines: we stopped tuning its optimization hyperparameters once it produced consistently good results on ImageNet.

The findings reveal that, even setting efficiency aside, current kernelized methods do not preserve the accuracy of pretrained softmax models they aim to replace. We conjecture that either kernelization is fundamentally limited as an approximation framework, or prior design choices (learned MLPs, frequency-domain transforms) leave room for improvement. The next section addresses this directly.

\section{LARA++: closing the quality gap}
\label{sec:lara}

To test whether the quality gap from \cref{sec:quality} is intrinsic to kernelization, we develop LARA++, an improved version of LARA. We chose LARA as the starting point because, among the three methods evaluated, it is the most principled: while Hedgehog and DiJiang make ad-hoc modifications to the feature map, LARA reformulates attention as an expectation under a Gaussian-mixture target distribution and approximates it via importance sampling. If any method has the structural foundations to support a quality-preserving extension, LARA does.


LARA reformulates attention computation as an expectation over random features. For a random feature map $\xi(\mathbf{x}, \omega) = \exp(\omega^{\top}\mathbf{x} - \frac{1}{2}\|\mathbf{x}\|^2)$, softmax attention equals:
\begin{equation}
\text{SoftmaxAttn}(\mathbf{q}_n, \mathbf{K}, \mathbf{V}) = \mathbb{E}_{\omega \sim p_n(\omega)} [f_n(\omega)],
\end{equation}
where $p_n(\omega) = \sum_{i=1}^{N} \pi_i \mathcal{N}(\omega; \mathbf{q}_n + \mathbf{k}_i, \mathbf{I})$ is a Gaussian mixture with weights $\pi_i \propto \exp(\mathbf{q}_n^{\top}\mathbf{k}_i)$, and:
\begin{equation}
f_n(\omega) = \frac{\xi(\mathbf{q}_n, \omega)^{\top} \sum_{i=1}^{N} \xi(\mathbf{k}_i, \omega)\mathbf{v}_i^{\top}}{\xi(\mathbf{q}_n, \omega)^{\top} \sum_{i=1}^{N} \xi(\mathbf{k}_{i}, \omega)}.
\end{equation}
Since $p_n(\omega)$ contains $N$ components (one per key), direct sampling is intractable. LARA instead employs self-normalized importance sampling with a tractable proposal distribution $q(\omega)$ from which samples are drawn $\omega_s \sim q(\omega)$:
\begin{equation}
\mathbb{E}_{p_n}[f_n(\omega)] \approx \frac{\sum_{s=1}^{S} w_s f_n(\omega_s)}{\sum_{s=1}^{S} w_s}, \quad w_s = \frac{p_n(\omega_s)}{q(\omega_s)}.
\end{equation}
The approximation quality depends on how closely $q(\omega)$ matches the target $p_n(\omega)$. LARA constructs proposals by partitioning the sequence into $C$ contiguous chunks, computing the mean query within each chunk as a \textit{landmark}, and defining each proposal $q_c(\omega)$ as a Gaussian centered at a weighted combination of the landmark and keys. Queries assigned to the same landmark therefore share a proposal distribution, from which LARA samples features to approximate the attention expectation.

\paragraph{Improvement 1: deterministic feature maps.}
LARA samples $\omega_s \sim q(\omega)$ to estimate the attention
expectation, introducing variance that degrades approximation. Following recent evidence that deterministic feature maps can match or exceed randomized ones~\citep{zhang2024hedgehog}, we eliminate sampling: we use the proposal-distribution means directly as deterministic features. \cref{sec:lara-ablation} confirms this consistently outperforms the randomized variant, on each of three independent seeds (\cref{tab:seeds}).

\paragraph{Improvement 2: clustering-based proposals.}
Contiguous chunking assumes that nearby queries have similar attention patterns, which need not hold: semantically related patches may be spatially distant in vision, and language models often rely on long-range dependencies. If queries within a chunk attend to different keys, their average may poorly represent any of them, increasing importance-sampling variance. We therefore replace contiguous chunks with adaptive clusters that group queries by similarity rather than position.
\begin{wrapfigure}[45]{r}{0.4\textwidth}
    \centering
    \includegraphics[width=\linewidth]{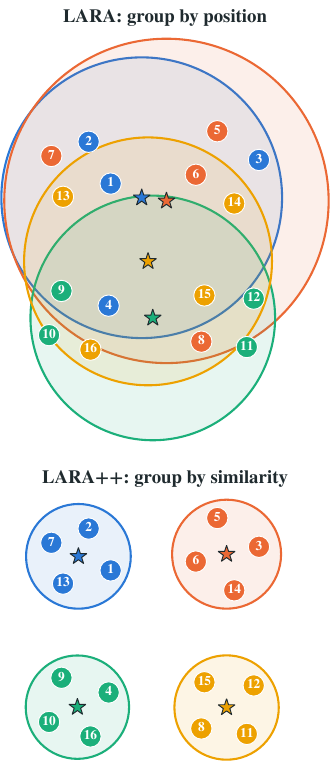}

    \caption{\textbf{Standard LARA (top) vs.\ LARA++ (bottom).} Same 16 query embeddings in both panels; colors mark proposal groups and $\star$ the landmark (group mean) defining each proposal. LARA groups contiguous positions, so each group mixes dissimilar queries and the landmarks collapse toward the global mean. LARA++ groups queries by similarity via greedy $k$-center clustering, giving tight groups with representative landmarks. The strip below each panel shows the group of each sequence position.}
\vspace{-1cm}
    \label{fig:lara_vs_lara_imp}
\end{wrapfigure}
While k-means is a natural choice, its iterative updates are too costly. Instead, we adapt the k-center greedy algorithm of \citet{sener2018active}. Starting from a random query, we repeatedly select the query farthest in cosine distance from the current centers until $C$ landmarks are chosen, then assign each query to its nearest center and average within each cluster. The procedure costs $O(NC)$, which is linear in $N$ only for fixed $C$. In practice, we set $C=N/k$ with $k\in\{8,16\}$, as this gives better accuracy than fixed $C$, consistent with the original LARA experiments. This yields $\Theta(N^2/k)$ clustering cost. The efficiency sweep in \cref{fig:lara} uses this regime; \cref{alg:kcenter} gives the algorithm, and \cref{fig:lara_vs_lara_imp} illustrates the difference.

\subsection*{Results on DeiT and DINOv2 distillation}

\paragraph{Distillation recipe.}
We convert pretrained softmax models to LARA++ through two-stage knowledge distillation. First, we sequentially replace each attention layer with LARA++ (starting from the first layer), training each replacement for one epoch while keeping the classification head frozen. The training objective combines a distillation loss, encouraging the linearized attention outputs to match the original across all layers, with a task-specific loss for prediction alignment. In the second stage, we fine-tune all parameters jointly  using only the task loss, allowing the model to adapt to the accumulated approximation error. For DeiT, the task loss combines feature cosine similarity with cross-entropy against the teacher's predictions. For DINOv2, we instead align CLS token outputs from the final layers and mean patch features to preserve self-supervised representations. Complete training details, loss formulations, and hyperparameters are provided in Appendix~\ref{app:exp-details}.
We ablate both improvements and the landmark count in Appendix~\ref{sec:lara-ablation}, confirming that each contributes to the final performance and that approximation quality improves with more landmarks.
For fair comparison, all baselines in \cref{sec:quality} were distilled within this same framework. Where the original authors provided distillation recipes, we preserved their method-specific customizations and tuned learning rates separately to each baseline's most stable configuration (see \cref{app:details-compared-methods}).

\paragraph{DeiT results.}
LARA++ consistently narrows the quality gap to softmax across all three DeiT configurations, reaching within $1.3$ point in each case and recovering $99\%$ of softmax accuracy on DeiT-B/384 ($82.1\%$ vs.\ $82.9\%$). Its gain over LARA is stable across model size and resolution ($0.8$--$1.1$ points), suggesting that the improvement is not tied to a single architectural regime. \cref{tab:matched-cost} shows the corresponding accuracy--efficiency trade-off: LARA++ uses $0.35$--$0.58\times$ the teacher's peak memory and outperforms LARA at equal memory, but is the slowest kernelized method, with DeiT-S training $1.2$--$1.3\times$ slower than the other baselines. This overhead is largely due to clustering, which accounts for $43$--$60\%$ of forward time on DINOv2 (\cref{app:lara-cost}). The same pattern holds on DINOv2, where LARA++ reduces peak memory by $1.8$--$2.5\times$ at a $0.9$--$2.3$ point ImageNet cost (\cref{tab:dino-matched}).

\begin{figure*}[t]
    \centering
    \includegraphics[width=0.995\linewidth]{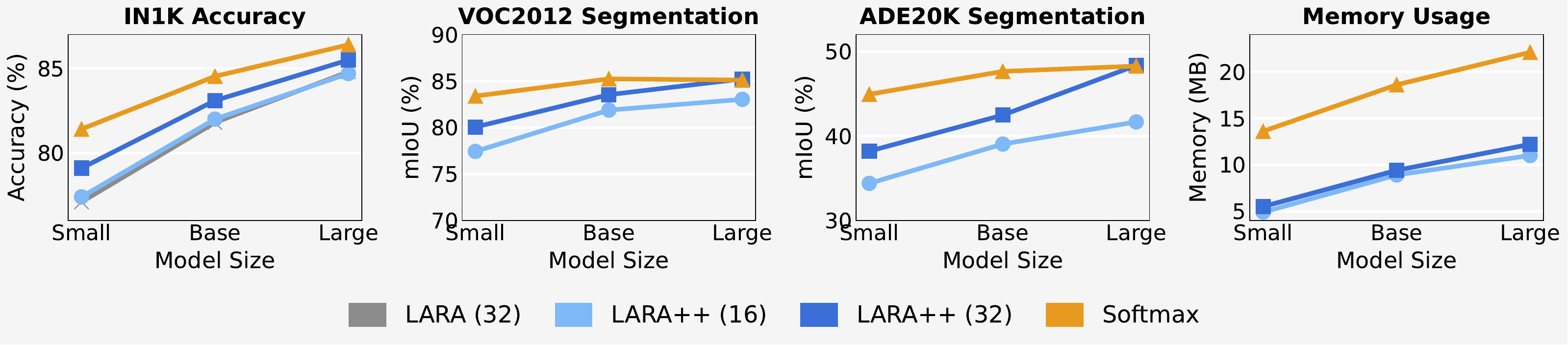}
    \caption{\textbf{DINOv2 results across model scales.} LARA++ (with 16 or 32 landmarks) closes the gap to softmax that standard LARA leaves open, and the gap narrows further with model capacity. On DINOv2-L, LARA++ (32) falls within $0.9$ points of softmax on ImageNet classification and matches it on both semantic segmentation tasks, while reducing memory usage by approximately 2$\times$. Memory includes the clustering stage; exact values are in \cref{tab:dino-matched}. DINOv2-g results are in \cref{tab:dino-giant}.}
    \label{fig:dino-experiments}
    \vspace{-0.25cm}
\end{figure*}

\paragraph{DINOv2 results.}
\cref{fig:dino-experiments} extends the evaluation to self-supervised pretraining and dense-prediction tasks across three model scales. As with DeiT, LARA leaves a consistent gap to softmax, while LARA++ narrows it across all three tasks. On DINOv2-Large, LARA++ is only $0.9$ points below softmax on ImageNet and matches softmax mIoU on VOC2012 and ADE20K. We further distill DINOv2-g (1.1B parameters, 40 layers), where the ImageNet gap shrinks to $0.5$ points and the VOC2012 gap to $0.7$ mIoU (\cref{tab:dino-giant}). The shrinking gap with scale mirrors the DeiT trend, while the dense-prediction results show LARA++'s gains extend beyond classification. See \cref{app:failure} for remaining failures.

\begin{wraptable}{r}{0.63\textwidth}
\vspace{-.5cm}
  \centering
\caption{Accuracy at matched cost on DeiT. Columns 2--4: top-1 (\%) / peak memory relative to the softmax teacher. Last column: DeiT-S/224 training step time; kernelized baselines use FLA kernels, and LARA++ includes clustering.}
\label{tab:matched-cost}
\small
\setlength{\tabcolsep}{4pt}
\begin{tabular}{lcccc}
\toprule
& DeiT-S/224 & DeiT-B/224 & DeiT-B/384 & train ms/step \\
\midrule
Softmax  & 79.8 / 1.00 & 81.8 / 1.00 & 82.9 / 1.00 & --- \\
Hedgehog & 77.7 / 0.41 & 79.7 / 0.71 & 80.4 / 0.52 & 294.6 \\
DiJiang  & 71.9 / 0.31 & 74.5 / 0.48 & 72.8 / 0.48 & 274.8 \\
LARA     & 77.4 / 0.35 & 80.0 / 0.49 & 81.0 / 0.58 & 289.6 \\
LARA++   & 78.5 / 0.35 & 80.8 / 0.49 & 82.1 / 0.58 & 352.1 \\
\bottomrule
\end{tabular}
\vspace{-.25cm}
\end{wraptable}
\paragraph{The two halves of the picture.}
LARA++ shows that the quality gap in \cref{sec:quality} is not fundamental: a principled approximation can recover near-softmax accuracy. The efficiency limitations of \cref{sec:efficiency}, however, remain. LARA++ inherits LARA's importance-sampling overhead and adds a clustering stage that dominates its forward time (\cref{tab:matched-cost}). As \cref{fig:lara} shows, even with aggressive landmark ratios, memory gains appear only at sequence lengths of $64$K and beyond, while time gains require even larger landmark counts. Thus, approximation quality can be recovered within the kernelized framework, but its promised practical efficiency gains remain elusive.

\section{Conclusion}

We asked whether the theoretical complexity advantages of kernelized attention translate into practical gains over exact attention. Across vision transformers and causal language models, our results show that they largely do not. FlashAttention matches or surpasses SOTA kernelized methods in training time and memory and throughout vision inference, despite its $O(N^2)$ compute complexity. The main exception is causal-LM inference, where recurrent linear attention provides substantially lower prefill and generation memory, with generation latency benefits in some settings. These results highlight that asymptotic complexity alone is a poor predictor of practical efficiency on modern hardware.

The quality results expose a second limitation. Existing kernelized methods remain $2$--$10$ points below softmax under a unified distillation protocol, and the gap does not disappear with model scale. LARA++ shows that this loss is not intrinsic to kernelization: improved proposal distributions and deterministic features recover near-softmax performance, reaching within one point on DeiT and similarly closing the gap across DINOv2 scales and downstream tasks. However, this improvement comes with substantial importance-sampling and clustering overhead, so solving the approximation problem does not solve the efficiency problem. 

Taken together, these findings suggest that the central challenge for efficient attention is no longer obtaining linear asymptotic complexity, but designing computations that map efficiently onto modern hardware while preserving softmax quality. For pure kernelized attention used as a drop-in replacement, our experiments find little practical advantage over hardware-optimized exact attention outside long causal inference. Hybrid approaches that combine linear global computation with selective exact attention may therefore offer a more promising direction than replacing softmax entirely.

\newpage
\section*{Acknowledgements}
 We acknowledge the support of the Natural Sciences and Engineering Research Council of Canada (NSERC) and the Digital Research Alliance of Canada (alliancecan.ca). Cem Subakan, and Mirco Ravanelli are respectively supported by the NSERC projects RGPIN-2023-05759, and RGPIN-2022-05088. Mirco Ravanelli also is supported by Translated through the Imminent research grant. Çağatay Yıldız is a member of the Machine Learning Cluster of Excellence, funded by the Deutsche Forschungsgemeinschaft (DFG, German Research Foundation) under Germany's Excellence Strategy – EXC number 2064/1 – Project number 390727645. This research utilized compute resources at the Tübingen Machine Learning Cloud, DFG FKZ INST 37/1057-1 FUGG. We have not received any support for this work except the regular research funding indicated above, and we declare no competing interests. We thank Jihoon Jeong for his helpful comments.
\bibliography{main}
\bibliographystyle{abbrvnat}

\appendix
\newpage

\section{LARA++ Ablations}
\label{sec:lara-ablation}

Table~\ref{tab:ablations} presents ablation studies on DeiT-small to validate our design choices. We compare four LARA variants: standard LARA (with contiguous chunking), LARA with random landmark selection, LARA++ with sampled (stochastic) feature maps, and our full LARA++ with clustering-based proposals and deterministic features.

The results confirm the effectiveness of both proposed improvements. Replacing contiguous chunking with random selection degrades accuracy from 77.4\% to 76.4\%, indicating that arbitrary partitioning performs worse than positional locality. Importantly, our clustering-based approach (LARA++) achieves 78.5\%, outperforming both alternatives by learning adaptive proposals that better cover the query distribution. The comparison between sampled (78.2\%) and deterministic (78.5\%) feature maps demonstrates that removing stochasticity provides a modest but consistent improvement. Because this $0.3$-point difference comes from a single run, we repeat both variants with three independent seeds (\cref{tab:seeds}). Deterministic features win on every seed (per-seed gaps $+0.3$, $+0.6$, $+0.5$; mean $78.33 \pm 0.15$ vs.\ $77.87 \pm 0.29$; paired $t$-test $p=0.034$). This supports the claim, although with $n=3$ the evidence is limited.

We also examine the effect of landmark count on approximation quality. Reducing landmarks from 28 to 20 decreases accuracy from 78.5\% to 78.0\%, while further reduction to 14 landmarks yields 77.5\%. This graceful degradation suggests that LARA++ can trade off accuracy for efficiency when computational budgets are constrained.

\begin{table}[h]
\caption{Ablation study on DeiT-small comparing LARA variants (single run, seed 0; see \cref{tab:seeds} for repeated runs).}
\label{tab:ablations}
  \centering
  \begin{tabular}{l|c|c}
    \toprule
    \textsc{Attention Type} & \textsc{\# Landmarks} & \textsc{Acc. (\%)}  \\
    \midrule
    Softmax & $28$ & $79.8$  \\
     LARA (chunking) & $28$ & $77.4$   \\
     LARA (chunking) & $14$ & $76.1$   \\
     
     LARA (random) & $28$ & $76.4$  \\
     LARA++ (sampled) & $28$ & $78.2$  \\
     LARA++ & $28$ & $78.5$   \\
     LARA++ & $20$ & $78.0$   \\
     LARA++ & $14$ & $77.5$  \\
    \bottomrule
  \end{tabular}
\end{table}

\begin{table}[h]
\centering

\caption{Repeated runs on DeiT-S/224 (top-1 \%, 28 landmarks for LARA variants). Mean $\pm$ sample standard deviation over three independent seeds.}
\label{tab:seeds}
\small
\begin{tabular}{lcccc}
\toprule
Method & Seed 0 & Seed 42 & Seed 23 & Mean $\pm$ Std \\
\midrule
Hedgehog & 77.7 & 77.6 & 77.6 & $77.63 \pm 0.06$ \\
DiJiang & 71.9 & 71.8 & 71.8 & $71.83 \pm 0.06$ \\
LARA & 77.4 & 77.5 & 77.2 & $77.37 \pm 0.15$ \\
\midrule
LARA++ (sampled) & 78.2 & 77.7 & 77.7 & $77.87 \pm 0.29$ \\
LARA++ (deterministic) & 78.5 & 78.3 & 78.2 & $78.33 \pm 0.15$ \\
\bottomrule
\end{tabular}
\end{table}

\begin{figure}
    \centering
    \includegraphics[width=0.995\linewidth]{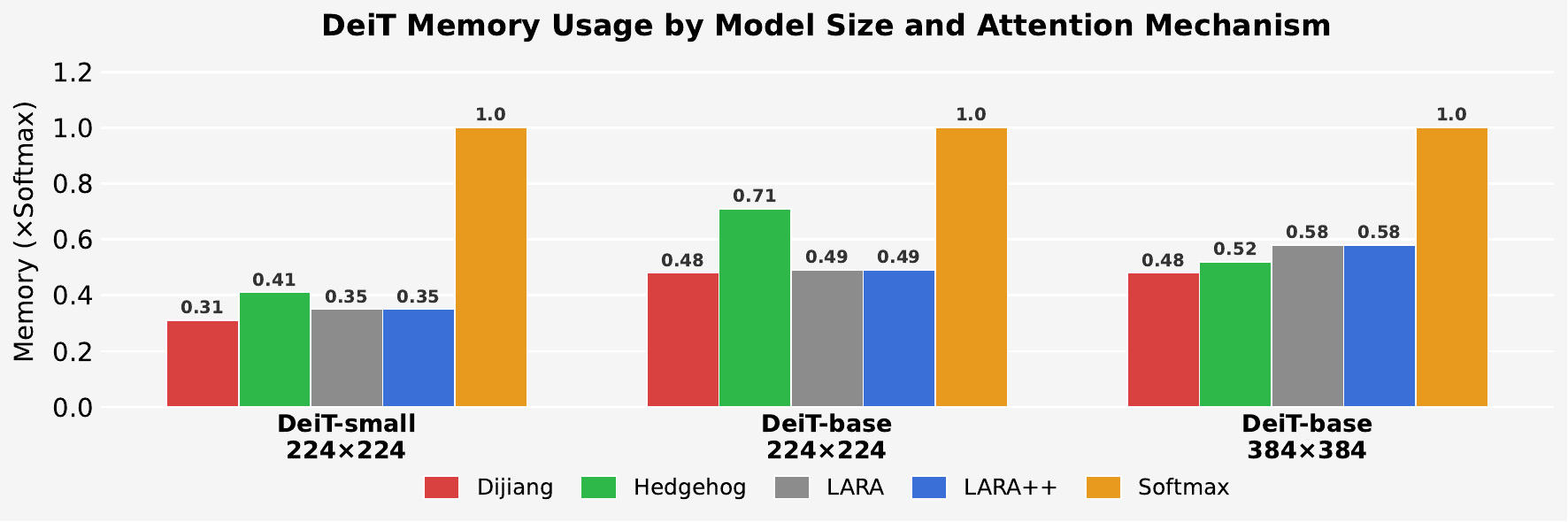}
    \caption{DeiT memory consumption with different model and input sizes.}
    \label{fig:deit-memory}
\end{figure}

\begin{algorithm}
\caption{\textit{K-Center Clustering for Query-Key Proposals}}
\label{alg:kcenter}
\begin{algorithmic}[1]
\REQUIRE Query $\mathbf{Q} \in \mathbb{R}^{L \times D}$, Key $\mathbf{K} \in \mathbb{R}^{L \times D}$, number of proposals $m$
\ENSURE Clustered proposals $\tilde{\mathbf{Q}}, \tilde{\mathbf{K}} \in \mathbb{R}^{m \times D}$
\STATE $\mathbf{Z} \gets \mathrm{Normalize}\left(\mathbf{Q} + \mathbf{K}\right)$
\STATE $\mathcal{C} \gets \{\text{initial index from } \{1, \dots, L\}\}$
\STATE $\mathbf{d}_i \gets d_{\cos}(\mathbf{z}_i, \mathbf{z}_{\mathcal{C}_1}), \quad \forall i$
\FOR{$j = 2$ \textbf{to} $m$}
    \STATE $c_j \gets \arg\max_i \, \mathbf{d}_i$; \quad $\mathcal{C} \gets \mathcal{C} \cup \{c_j\}$
    \STATE $\mathbf{d}_i \gets \min\left(\mathbf{d}_i, \, d_{\cos}(\mathbf{z}_i, \mathbf{z}_{c_j})\right), \quad \forall i$
\ENDFOR
\STATE $a_i \gets \arg\min_{c \in \mathcal{C}} \, d_{\cos}(\mathbf{z}_i, \mathbf{z}_c), \quad \forall i$
\FOR{$j = 1$ \textbf{to} $m$}
    \STATE $\mathcal{S}_j \gets \{i : a_i = j\}$; \quad $\tilde{\mathbf{q}}_j \gets \frac{1}{|\mathcal{S}_j|} \sum_{i \in \mathcal{S}_j} \mathbf{q}_i$; \quad $\tilde{\mathbf{k}}_j \gets \frac{1}{|\mathcal{S}_j|} \sum_{i \in \mathcal{S}_j} \mathbf{k}_i$
\ENDFOR
\STATE \textbf{return} $\tilde{\mathbf{Q}}, \tilde{\mathbf{K}}$
\end{algorithmic}
\end{algorithm}


\section{Efficiency results across model scales}
\label{app:scale}

\begin{figure*}[t]
    \centering
    \includegraphics[width=0.99\linewidth]{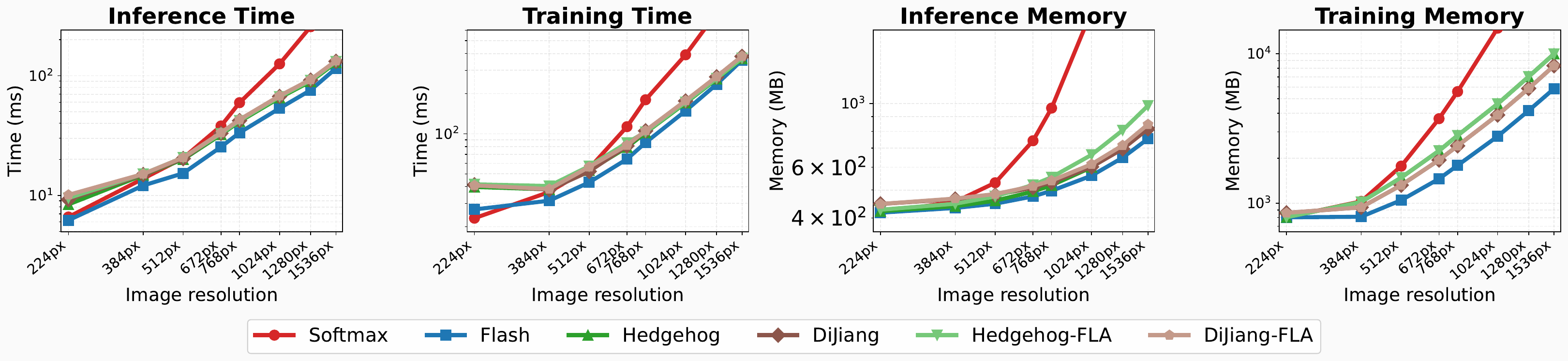}
    \includegraphics[width=0.99\linewidth]{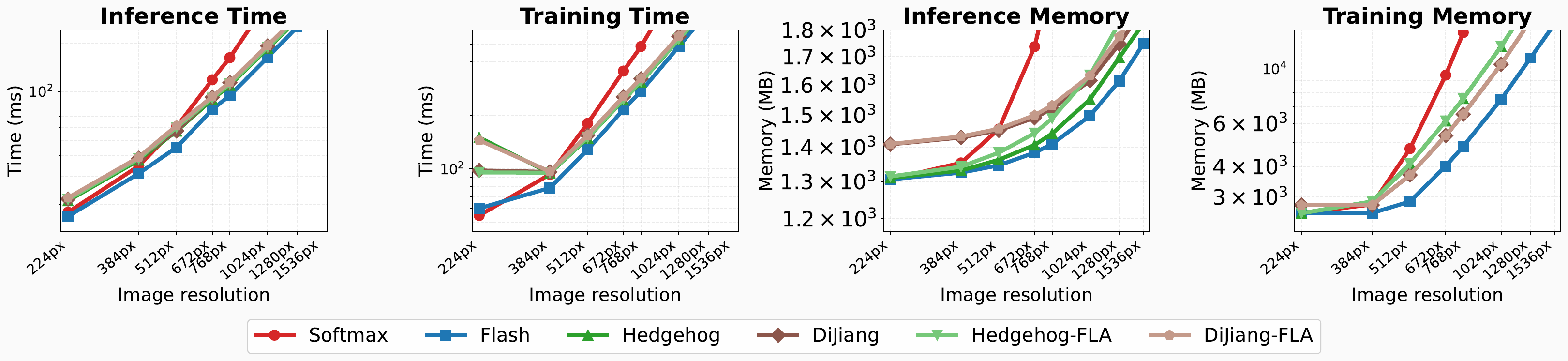}
    \caption{Memory and computational efficiency across sequence lengths for ViT-B (top) and ViT-L (bottom), A100 GPU. The ordering matches ViT-S (\cref{fig:efficiency}).}
    \label{fig:app:vit-scale}
\end{figure*}

\begin{figure*}[t]
    \centering
    \includegraphics[width=0.99\linewidth]{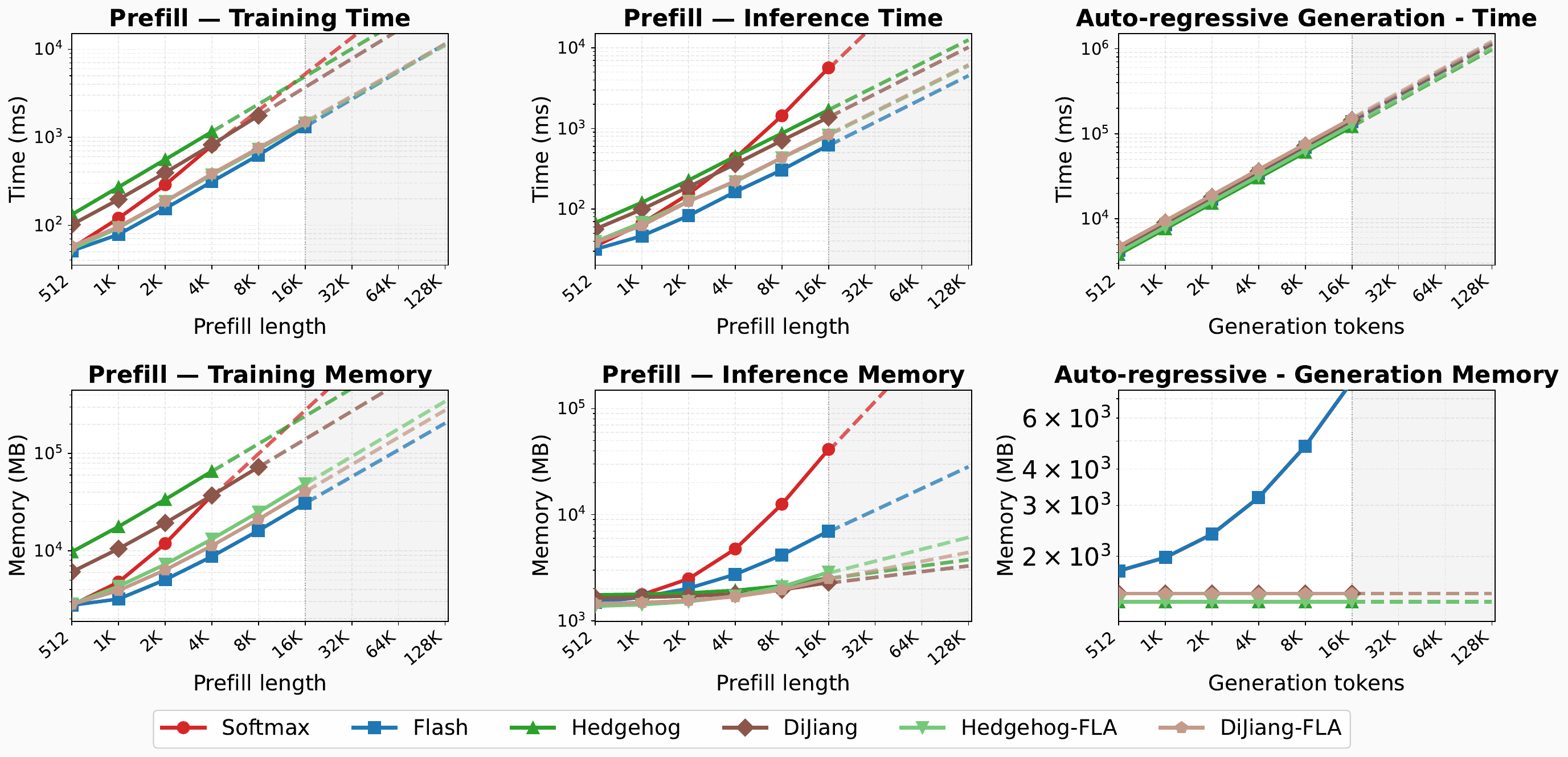}
    \includegraphics[width=0.99\linewidth]{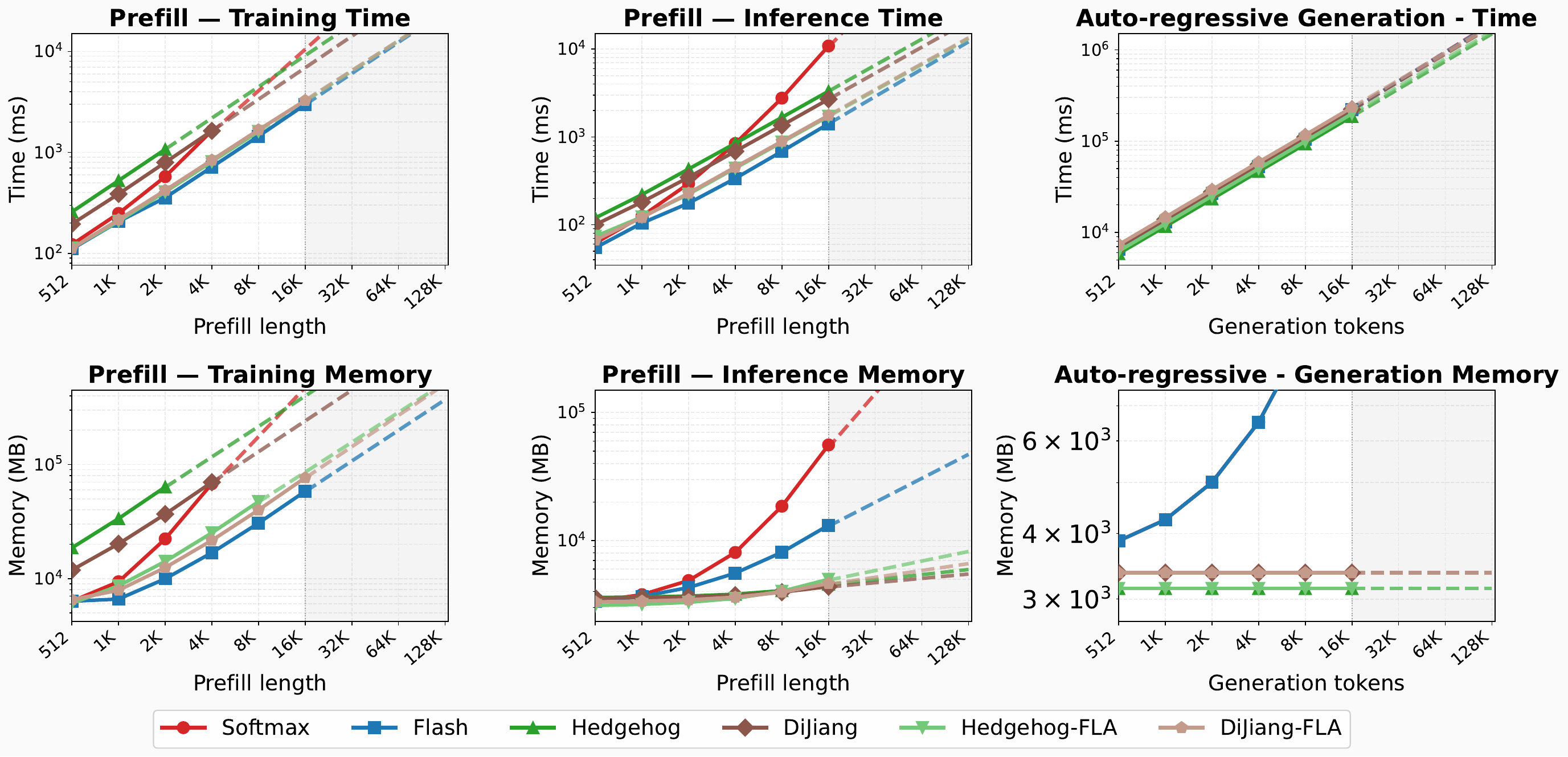}
    \caption{Memory and computational efficiency across sequence lengths for GPT2-M (top) and GPT2-L (bottom), A100 GPU. Markers and solid lines are measured; dashed lines are log-domain extrapolations.}
    \label{fig:app:gpt-scale}
\end{figure*}

{
This section reports the larger-model counterparts of \cref{fig:efficiency,fig:efficiency-causal}, using the same protocol: a single A100 GPU, 100 CUDA-synchronized timed runs, and peak memory from \texttt{max\_memory\_allocated}. \cref{tab:scale-vit} summarizes the vision transformers at the longest measured resolution (1536px, $N=9217$). \cref{tab:scale-gpt} summarizes the causal language models: training at $N=4$K (the longest length at which most methods still fit), prefill inference at $N=16$K, and generation of $M=16$K tokens from a 512-token prompt. The full sweeps are shown in \cref{fig:app:vit-scale,fig:app:gpt-scale}.
}

\begin{table}[h]
\centering

\caption{Vision transformers at 1536px ($N=9217$), time (ms) / peak memory (MB). FlashAttention is the fastest and lowest-memory method in every row.}
\label{tab:scale-vit}
\small
\setlength{\tabcolsep}{3.5pt}
\resizebox{\linewidth}{!}{%
\begin{tabular}{lcccccc}
\toprule
& Softmax & Flash & Hedgehog & DiJiang & Hedgehog-FLA & DiJiang-FLA \\
\midrule
ViT-S inf   & 269.7 / 4294  & \textbf{69.0 / 302}    & 82.5 / 347     & 82.0 / 327     & 82.2 / 421     & 82.4 / 342 \\
ViT-B inf   & 499.5 / 8740  & \textbf{115.2 / 754}   & 128.7 / 841    & 131.7 / 816    & 131.3 / 983    & 133.2 / 846 \\
ViT-L inf   & 1351.5 / 12398 & \textbf{363.7 / 1748} & 390.0 / 1826   & 402.9 / 1887   & 396.5 / 2015   & 407.0 / 1926 \\
\midrule
ViT-S train & 1205.7 / 33341 & \textbf{144.9 / 2830}  & 156.6 / 4935   & 157.0 / 4085   & 158.5 / 4960   & 158.4 / 4085 \\
ViT-B train & 2445.3 / 66843 & \textbf{357.4 / 5844}  & 366.4 / 9996   & 381.0 / 8323   & 369.7 / 9972   & 382.6 / 8323 \\
ViT-L train & OOM            & \textbf{1134.0 / 15261} & 1143.7 / 26162 & 1194.6 / 21727 & 1150.4 / 26162 & 1199.0 / 21727 \\
\bottomrule
\end{tabular}}
\end{table}

\begin{table}[h]
\centering

\caption{Causal language models, time (ms) / peak memory (MB). Bold marks the best time and the best memory in each row.}
\label{tab:scale-gpt}
\small
\setlength{\tabcolsep}{2.5pt}
\resizebox{\linewidth}{!}{%
\begin{tabular}{lcccccc}
\toprule
& Softmax & Flash & Hedgehog & DiJiang & Hedgehog-FLA & DiJiang-FLA \\
\midrule
GPT2-S train @4K    & 315.7 / 15051   & \textbf{111.6 / 3264}   & 633.7 / 24682   & 392.6 / 13867   & 130.6 / 4930    & 131.0 / 4201 \\
GPT2-M train @4K    & 801.7 / 37067   & \textbf{312.1 / 8733}   & 1152.8 / 65681  & 820.7 / 37033   & 373.2 / 13175   & 383.6 / 11257 \\
GPT2-L train @4K    & 1643.3 / 67973  & \textbf{716.6 / 16850}  & OOM             & 1635.6 / 69961  & 808.9 / 25179   & 838.7 / 21630 \\
\midrule
GPT2-S prefill @16K & 2961.9 / 28680  & \textbf{215.2} / 2886   & 804.2 / 1325    & 549.6 / \textbf{1106}  & 304.0 / 1583 & 304.8 / 1256 \\
GPT2-M prefill @16K & 5645.2 / 41239  & \textbf{621.0} / 7014   & 1704.0 / 2534   & 1373.2 / \textbf{2291} & 820.6 / 2866 & 839.8 / 2491 \\
GPT2-L prefill @16K & 10872.8 / 55828 & \textbf{1413.9} / 13180 & 3321.7 / 4569   & 2691.6 / \textbf{4372} & 1719.8 / 4984 & 1761.0 / 4621 \\
\midrule
GPT2-S gen @16K     & 82614 / 3016    & 90041 / 2991    & \textbf{79514} / 533   & 93446 / 560    & 90952 / \textbf{481}  & 103023 / 507 \\
GPT2-M gen @16K     & 137542 / 8026   & 139415 / 8026   & \textbf{121467 / 1394} & 144445 / 1488  & 127926 / \textbf{1394} & 152114 / 1488 \\
GPT2-L gen @16K     & 219923 / 15576  & 221969 / 15576  & \textbf{186529 / 3144} & 222955 / 3368  & 199232 / \textbf{3144} & 234094 / 3368 \\
\bottomrule
\end{tabular}}
\end{table}

\newpage
\section{Cost of LARA++}
\label{app:lara-cost}

{
\cref{tab:dino-matched} gives the DINOv2 counterpart of \cref{tab:matched-cost}, and \cref{tab:deit-wallclock} gives the full DeiT-S per-step measurements. The \texttt{forward()} columns time the forward function of the full model. \cref{tab:clustering} isolates the clustering stage on DINOv2. Clustering accounts for $43$--$60\%$ of the LARA++ forward time. \cref{tab:hw-cost} profiles the hardware-level cost on DINOv2-L ($N=257$, batch size 8). LARA++ performs \emph{fewer} FLOPs than softmax ($0.94\times$ on DINOv2-L and $0.86$--$0.96\times$ across DINOv2-S/B/L/g), yet it launches ${\sim}340$ kernels versus softmax's 3 and moves ${\sim}12\times$ more DRAM traffic. Its cost is therefore bound by memory traffic and kernel launches, not by arithmetic. LARA and LARA++ share the same attention body and differ in landmark generation (and in replacing sampled features with deterministic ones), so the LARA$\rightarrow$LARA++ difference is attributable to clustering. Clustering is responsible for ${\sim}83\%$ of kernel launches and ${\sim}68\%$ of DRAM traffic at 32 landmarks (${\sim}72\%$ and ${\sim}60\%$ at 16), while contributing about $2\%$ of the analytic FLOPs. Halving the landmark budget from 32 to 16 removes about a third of the clustering kernels and cuts DRAM traffic by ${\sim}35\%$. At these sequence lengths, the residual overhead is therefore a clustering-and-implementation cost: a sequence of small gather/argmax/scatter kernels that do not fuse. Note, however, that with $C=N/k$ landmarks the clustering FLOPs grow quadratically with $N$ (\cref{sec:lara}).
}

\begin{table}[h]
\centering

\caption{DINOv2: ImageNet top-1 (\%) / peak memory (MB), including clustering.}
\label{tab:dino-matched}
\small
\begin{tabular}{lcccc}
\toprule
Model & Softmax & LARA (32) & LARA++ (16) & LARA++ (32) \\
\midrule
DINOv2-S & 81.4 / 13.6 & 77.1 / 5.5  & 77.4 / 4.9  & 79.1 / 5.5 \\
DINOv2-B & 84.5 / 18.6 & 81.8 / 9.4  & 82.0 / 8.9  & 83.1 / 9.4 \\
DINOv2-L & 86.4 / 22.1 & 84.8 / 12.2 & 84.7 / 11.0 & 85.5 / 12.2 \\
\bottomrule
\end{tabular}
\end{table}

\begin{table}[h]
\centering

\caption{DeiT-S per-step training cost and forward-pass cost.}
\label{tab:deit-wallclock}
\small
\begin{tabular}{lcccc}
\toprule
Method & ms/step & train MB & \texttt{forward()} ms & \texttt{forward()} MB \\
\midrule
Hedgehog-FLA & 294.6 & 19315 & 85.23  & 3331 \\
DiJiang-FLA  & 274.8 & 16390 & 68.87  & 2958 \\
LARA         & 289.6 & 15961 & 75.15  & 3313 \\
LARA++       & 352.1 & 16106 & 147.81 & 3330 \\
\bottomrule
\end{tabular}
\end{table}

\begin{table}[h]
\centering

\caption{Clustering cost of LARA++ on DINOv2 (forward pass).}
\label{tab:clustering}
\small
\begin{tabular}{lcccc}
\toprule
Model & LARA fwd (no clustering) & LARA++ fwd & Clustering (isolated) & Clustering share \\
\midrule
DINOv2-S & 3.6ms  & 8.9ms  & 5.4ms  & 60\% \\
DINOv2-B & 9.8ms  & 17.2ms & 8.4ms  & 49\% \\
DINOv2-L & 14.1ms & 24.5ms & 10.6ms & 43\% \\
DINOv2-g & 9.9ms  & 24.5ms & 14.5ms & 59\% \\
\bottomrule
\end{tabular}
\end{table}

\begin{table}[h]
\centering

\caption{Hardware-level cost on DINOv2-L ($N=257$, batch size 8, A100).}
\label{tab:hw-cost}
\small
\begin{tabular}{lccc}
\toprule
Method & Forward FLOPs (rel.\ softmax) & Kernel launches & DRAM traffic proxy (MB) \\
\midrule
Softmax      & 77.6G ($1.00\times$) & 3   & 168 \\
LARA (32)    & 71.3G ($0.92\times$) & 62  & 676 \\
LARA++ (32)  & 72.9G ($0.94\times$) & 344 & 2109 \\
\bottomrule
\end{tabular}
\end{table}


\section{Where does LARA++ still fail?}
\label{app:failure}

{
\cref{tab:failure} collects the DINOv2 results of \cref{fig:dino-experiments} for softmax and LARA++ with 32 and 16 landmarks. LARA++'s residual gap shrinks with both model capacity and landmark budget, and three failure modes are visible.

\textbf{ADE20K segmentation.} ADE20K (150 classes) is by far the hardest task: LARA++ (32) trails softmax by $6.7$ mIoU on Small and $5.2$ on Base, versus $2.3$ and $1.4$ points on ImageNet. The gap is not explained by dense prediction alone. On VOC2012 (21 classes) the gaps ($3.3$ and $1.7$) are close to those on ImageNet, so the failure concentrates on fine-grained, many-class scene parsing.

\textbf{Insufficient landmark budget.} Halving the landmarks from 32 to 16 costs $0.8$--$1.7$ points on ImageNet but $3.4$--$6.7$ mIoU on ADE20K. The damage is largest for the biggest model, where ADE20K drops from $48.4$ to $41.7$. When the representation has fine-grained structure to preserve, too few proposals destroy it, and the degradation is not graceful.

\textbf{Small models.} With 32 landmarks, DINOv2-L LARA++ matches softmax on both segmentation tasks ($85.2$ vs.\ $85.1$ on VOC2012, $48.4$ vs.\ $48.3$ on ADE20K), whereas DINOv2-S trails by $3.3$ and $6.7$. Standard LARA shows the same trend on DINOv2 (ImageNet gap $4.3 \rightarrow 2.7 \rightarrow 1.6$ from Small to Large), so larger models appear to tolerate importance-sampling approximations better in this self-supervised setting generally. This contrasts with the DeiT baselines in \cref{tab:deit-quality}, whose gaps do not close with scale, but the two settings differ in distillation objective and model family.

\textbf{Untested regimes.} We do not claim results for two regimes. First, LARA++ is evaluated on vision only, not on causal language modelling. Second, we have not tested very long contexts. There, a fixed landmark budget must summarise far more tokens per proposal, while a fixed landmark ratio makes clustering quadratic in $N$ (\cref{sec:lara}).
}

\begin{table}[h]
\centering
\caption{DINOv2 results per task: softmax / LARA++ (32 landmarks) / LARA++ (16 landmarks).}
\label{tab:failure}
\small
\begin{tabular}{lccc}
\toprule
Task & DINOv2-S & DINOv2-B & DINOv2-L \\
\midrule
IN1K top-1   & 81.4 / 79.1 / 77.4 & 84.5 / 83.1 / 82.0 & 86.4 / 85.5 / 84.7 \\
VOC2012 mIoU & 83.4 / 80.1 / 77.4 & 85.2 / 83.5 / 81.9 & 85.1 / 85.2 / 83.0 \\
ADE20K mIoU  & 44.9 / 38.2 / 34.4 & 47.7 / 42.5 / 39.1 & 48.3 / 48.4 / 41.7 \\
\bottomrule
\end{tabular}
\end{table}

\section{Implementation Details}
\label{app:exp-details}

\subsection{Distilling from Softmax to LARA++ Networks}

\paragraph{Training Details and Evaluations.}
For DeiT tasks, we conduct experiments with \textit{DeiT-small}, \textit{DeiT-base}, and \textit{DeiT-base-384}. All three models have 12 layers. The distillation procedure consists of two stages. In stage 1, each attention block is replaced with linear attention (starting from the first layer), keeping the pretrained query, key, value, and projection networks. Each replaced block is trained for a single epoch with the linear head frozen. In stage 2, all parameters are finetuned for five epochs. Hyperparameters are in Table~\ref{details:deit}. We define:
\begin{itemize}[itemsep=0pt,topsep=2pt]
\item $o_{i}$: original model's $i^{\text{th}}$ layer attention output; \quad $l_{i}$: linearized model's $i^{\text{th}}$ layer attention output.
\item $f_o$/$f_l$: original/linearized model's features before linear classifier; \quad $\text{logits}_l$: linearized model's logits.
\item $\hat{y}_o$: original model's predictions.
\end{itemize}
We utilize the loss from \citet{ranzinger2024amradio} as the distillation loss, with $\alpha=0.9$ and $\beta=0.1$:
\begin{align}
L_{\text{distillation}}(\mathbf{O}, \mathbf{L}) &= \sum_{i=1}^{N} \alpha L_{\cos}(o_i, l_i) + \beta L_{\text{smooth-L1}}(o_i, l_i) \\
L_{\text{task}} &= L_{\cos}(f_o, f_l) + L_{\text{CE}}(\hat{y}_o, \text{logits}_l) \\
L_{\text{DeiTStage}_1} &= L_{\text{distillation}} + L_{\text{task}}; \quad L_{\text{DeiTStage}_2} = L_{\text{task}}
\end{align}

For DINOv2 tasks, we conduct experiments with \textit{Small}, \textit{Base}, \textit{Large}, and \textit{Giant} models (12, 12, 24, and 40 layers respectively), using a two-stage strategy with modifications to $L_{\text{task}}$. Since YFCC15M is very large, we only optimize for 2\,500 steps per epoch. Details in Table~\ref{details:dino}. Defining $\text{CLS}_{o,i}$/$\text{CLS}_{l,i}$ as original/linearized $i^{\text{th}}$ layer CLS outputs and $\bar{f}_o$/$\bar{f}_l$ as mean output features (excluding CLS):
\begin{align}
L_{\text{taskDino}} &= \sum_{i=N-3}^{N} L_{\cos}(\text{CLS}_{o,i}, \text{CLS}_{l,i}) + L_{\cos}(\bar{f}_o, \bar{f}_l) \\
L_{\text{DinoStage}_1} &= L_{\text{distillation}} + L_{\text{taskDino}}; \quad L_{\text{DinoStage}_2} = L_{\text{taskDino}}
\end{align}

For semantic segmentation downstream tasks, we finetune only the segmentation head: $L_{\text{DinoSemSeg}} = L_{\text{CE}}(\hat{y}_{\text{distilled}}, y_{\text{mask}})$. Details in Table~\ref{details:semseg}.

\begin{table}[h]
\centering
\caption{Hyperparameters of DeiT experiments. DeiT-base-384 uses $0.25$ of training data per distillation epoch.}
\label{details:deit}
\small
\begin{tabular}{lccccccc}
\toprule
\textbf{Model} & \textbf{Dataset} & \textbf{BS} & \textbf{Dist. LR} & \textbf{\# Dist. Ep.} & \textbf{FT LR} & \textbf{\# FT Ep.} \\
\midrule
DeiT-small & ImageNet-1K & 256 & 3e-5 & 1$\times$12 & 2e-6 & 5 \\
DeiT-base & ImageNet-1K & 256 & 3e-5 & 1$\times$12 & 2e-6 & 5 \\
DeiT-base-384 & ImageNet-1K & 64 & 2e-5 & 0.25$\times$12 & 2e-6 & 3 \\
\bottomrule
\end{tabular}
\end{table}

\begin{table}[h]
\caption{Hyperparameters of DINOv2 experiments. A limited amount of data is used.}
\centering
\label{details:dino}
\small
\begin{tabular}{lcccccc}
\toprule
\textbf{Model} & \textbf{Dataset} & \textbf{BS} & \textbf{Dist. LR} & \textbf{\# Dist. Steps} & \textbf{FT LR} & \textbf{\# FT Steps} \\
\midrule
Small & YFCC15M & 32 & 2e-5 & 2500$\times$12 & 3e-6 & 2500$\times$8 \\
Base & YFCC15M & 32 & 2e-5 & 2500$\times$12 & 3e-6 & 2500$\times$8 \\
Large & YFCC15M & 32 & 2e-5 & 2500$\times$12 & 3e-6 & 2500$\times$8 \\
\bottomrule
\end{tabular}
\end{table}

\begin{table}[h]
\centering
\caption{Semantic segmentation finetuning details. VOC2012: 2\,000 training images; ADE20K: 20\,000.}
\label{details:semseg}
\small
\begin{tabular}{lcccc}
\toprule
\textbf{Model} & \textbf{Dataset} & \textbf{BS} & \textbf{FT LR} & \textbf{\# Epochs} \\
\midrule
Small/Base/Large & ADE20k & 16 & 3e-4 & 2 \\
Small/Base/Large & VOC2012 & 16 & 5e-4 & 10 \\
\bottomrule
\end{tabular}
\end{table}

\subsection{Details about the Comparison Partners}
\label{app:details-compared-methods}

For Hedgehog, we follow their recipe closely but tune the learning rate as the pipeline collapses with original settings. Stage 1 trains only $\phi_{\text{query}}$ and $\phi_{\text{key}}$ for two epochs:
\begin{equation}
L_{\text{HedgehogStage1}} = L_{\text{CE}}\!\left(\text{softmax}\!\left(\tfrac{QK^T}{\sqrt{d_k}}\right), \tfrac{\phi_{\text{query}}(Q)\phi_{\text{key}}(K)^T}{\sum(\phi_{\text{query}}(Q)\phi_{\text{key}}(K)^T)}\right)
\end{equation}
Stage 2 finetunes all parameters with cross-entropy. The original second-stage LR (0.001) causes collapse; we lower it to 0.000002. Details in Table~\ref{details:hedgehog}.

\begin{table}[h]
\centering
\caption{Hyperparameters of Hedgehog DeiT experiments.}
\label{details:hedgehog}
\small
\begin{tabular}{lcccccc}
\toprule
\textbf{Model} & \textbf{Dataset} & \textbf{BS} & \textbf{Dist. LR} & \textbf{\# Dist. Ep.} & \textbf{FT LR} & \textbf{\# FT Ep.} \\
\midrule
DeiT-small & ImageNet-1K & 32 & 0.02 & 2 & 2e-6 & 8 \\
DeiT-base & ImageNet-1K & 32 & 0.02 & 2 & 2e-6 & 8 \\
DeiT-base-384 & ImageNet-1K & 16 & 0.02 & 0.25$\times$2 & 2e-6 & 0.25$\times$8 \\
\bottomrule
\end{tabular}
\end{table}

For DiJiang, we were unable to make it work with their proposed scheme. We initialize scaling parameters $D_1$ and $D_2$ as 1 to avoid collapse, and employ two-stage training: stage 1 trains $D_1$, $D_2$, and gating $W_g$; stage 2 optimizes all parameters. Details in Table~\ref{details:djalo}.

\begin{table}[h]
\centering
\caption{Hyperparameters of DiJiang DeiT experiments.}
\label{details:djalo}
\small
\begin{tabular}{lcccccc}
\toprule
\textbf{Model} & \textbf{Dataset} & \textbf{BS} & \textbf{Dist. LR} & \textbf{\# Dist. Ep.} & \textbf{FT LR} & \textbf{\# FT Ep.} \\
\midrule
DeiT-small & ImageNet-1K & 256 & 0.001 & 2 & 5e-5 & 8 \\
DeiT-base & ImageNet-1K & 256 & 0.001 & 2 & 5e-5 & 8 \\
DeiT-base-384 & ImageNet-1K & 64 & 0.001 & 0.25$\times$2 & 5e-5 & 0.25$\times$8 \\
\bottomrule
\end{tabular}
\end{table}

\begin{figure*}[t]
    \centering
    \includegraphics[width=0.95\linewidth]{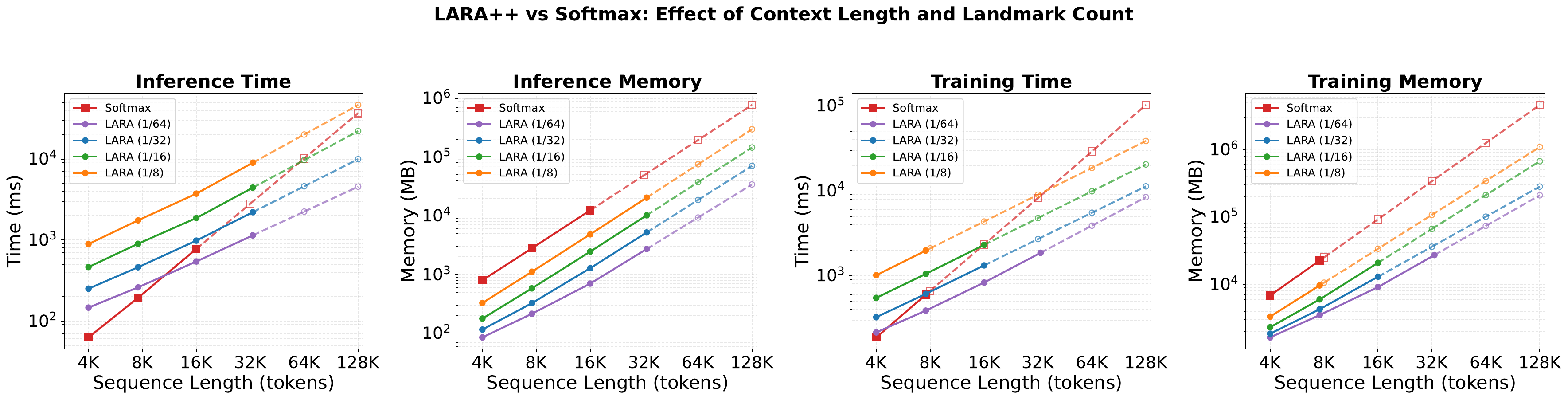}
    \caption{\textbf{Effect of landmark ratio on LARA++ efficiency.} Number of landmarks (as fractions of sequence length: 1/64, 1/32, 1/16, 1/8) vs.\ LARA++ memory and computational time. Lower landmark ratios reduce memory but increase computation time. At 64K+, LARA++ achieves memory savings across all configurations, though time advantage only materializes for higher landmark ratios. All timings are full forward passes including the clustering stage. The legend label ``LARA'' refers to LARA++. Because the number of landmarks grows with $N$, the $O(NC)$ clustering stage (\cref{alg:kcenter}) is quadratic in $N$ in this sweep.}
    \label{fig:lara}
\end{figure*}

\begin{table}[t]
\centering

\caption{DINOv2-g (1.1B parameters) distilled to LARA++ with 32 landmarks (ratio $\approx 1/8$).}
\label{tab:dino-giant}
\small
\begin{tabular}{lccc}
\toprule
& Softmax & LARA++ (32) & $\Delta$ \\
\midrule
IN1K top-1 & 86.5 & 86.0 & $-0.5$ \\
VOC2012 mIoU & 86.70 & 85.99 & $-0.71$ \\
\bottomrule
\end{tabular}
\end{table}


\end{document}